\documentclass{article}
\usepackage{graphicx} % Required for inserting images

\usepackage{MM805}
\usepackage{float}
\title{Image Recognition for Garbage Classification Based on Pixel Distribution Learning}
\author{Jenil Kanani}
\affil{Department of Computing Science, University of Alberta}
\affil{\tt{jenilhar@ualberta.ca}}
\date{July 16, 2024}

\begin{document}

\maketitle

\section{Abstract}
The exponential growth in waste production due to rapid economic and industrial development necessitates efficient waste management strategies to mitigate environmental pollution and resource depletion. Leveraging advancements in computer vision, this study proposes a novel approach inspired by pixel distribution learning techniques to enhance automated garbage classification. The method aims to address limitations of conventional convolutional neural network (CNN)-based approaches, including computational complexity and vulnerability to image variations. We will conduct experiments using the Kaggle Garbage Classification dataset, comparing our approach with existing models to demonstrate the strength and efficiency of pixel distribution learning in automated garbage classification technologies.

\section{Introduction}
The rapid economic and industrial development has led to an exponential increase in the volume of waste generated daily. Failure to effectively manage this increasing waste stream not only precipitates severe environmental pollution \cite{8807233, ANVARIFAR2018144} but also results in a waste of valuable resources. Recognizing the urgent need for efficient waste management solutions, governments are increasingly turning their focus towards recycling strategies as a means to mitigate environmental impact and foster sustainable development. Central to these efforts is the imperative to implement robust sorting, recycling, and regeneration processes.

Manual garbage sorting, while yielding highly accurate results, is labor-intensive and reliant on well-trained operators, thereby impeding overall efficiency \cite{su14053099}. Consequently, there arises a pressing need for automated garbage sorting solutions. The advancement of artificial intelligence (AI) technology has demonstrated remarkable potential in AI-related applications across numerous industries, particularly in computer vision (CV). Notably, AI-driven approaches have been increasingly explored in the domain of garbage classification tasks.

Historically, Support Vector Machines (SVMs) emerged as a popular supervised learning method for garbage classification \cite{yang2016classification}. However, the rapid evolution of deep learning has prompted a shift towards convolutional neural network (CNN)-based methods, offering improved accuracy in image classification tasks \cite{meng2020study}. Nonetheless, as CNN architectures become more complex, they demand escalating computational resources for training \cite{han2018advanced}. Additionally, conventional convolutional approaches often exhibit vulnerability to factors like color shifts and affine transformations, which means that supplementary data augmentation procedures are needed to overcome such limitations \cite{hernandez2018further}.

Against this backdrop, this project draws inspiration from recent advancements in pixel distribution learning techniques employed in background subtraction tasks \cite{zhao2019dynamic}. We propose a novel approach using pixel distribution learning, designed to address the inherent limitations of CNN-based approaches in image classification. By leveraging insights from pixel distribution learning, the proposed method aims to retain classification accuracy while overcoming the computational overhead and robustness issues encountered by conventional CNN-based approaches.

To validate the efficacy of the proposed approach, we will conduct experiments using the Garbage Classification dataset on Kaggle \cite{kaggle_garbage_classification}, comparing the performance of our method against existing models. Our objective in this project is to make a meaningful contribution to the advancement of automated garbage classification technologies, thereby providing more efficient garbage classification practices based on image classification.

\subsection{Related Work}
In the initial phase of image classification, conventional supervised techniques such as Support Vector Machines (SVM) and k-Nearest Neighbors (KNN) were predominantly used. Y. Lin et al. employed SVM for large-scale image classification, yielding commendable accuracy rates \cite{5995477}. Similarly, M. Pal et al. conducted a comparative analysis of SVM-based methodologies, including Relevance Vector Machine (RVM) and Sparse Multinomial Logistic Regression (SMLR), to evaluate their performance \cite{6331573}.

With the development of the hardware computation capability, the domain of image classification has witnessed significant advancements with the advent of deep learning techniques. Two primary streams of research have emerged, focusing on convolutional neural networks (CNNs) and transformer-based models, each demonstrating unique strengths in handling image data.

\textbf{CNN-based Models:} Among the CNN architectures, ResNet-32, a variant of the Residual Networks, has garnered attention for its ability to mitigate the vanishing gradient problem through skip connections. This design enables the training of deeper networks, which is crucial for complex image classification tasks. ResNet-32 has shown remarkable performance in various benchmarks and is recognized for its efficiency and accuracy in processing image data \cite{he2016deep}.

\textbf{Vision Transformer (ViT):} The introduction of Vision Transformers marks a significant paradigm shift in image classification. Unlike traditional CNNs that process images through localized convolutional filters, ViT employs a self-attention mechanism that allows it to consider the entire image at once. This global perspective enables ViT to capture long-range interactions between different parts of the image more effectively than CNNs. As a result, ViT can better understand complex spatial relationships within the image, leading to improved classification accuracy, especially in scenarios where context or relation between distant image parts is crucial. Furthermore, ViT's scalability and performance improve significantly with the increase in data and model size, showcasing its potential in leveraging large datasets more effectively than traditional CNNs \cite{dosovitskiy2020image}

\textbf{OpenAI CLIP:} CLIP (Contrastive Language-Image Pretraining) from OpenAI is notable not only for its integration of natural language processing and computer vision but also for its remarkable zero-shot learning capabilities. Unlike traditional CNN models that require extensive training on task-specific datasets, CLIP can accurately classify images it has never seen during training. This is achieved by leveraging a large-scale dataset of images and textual descriptions, allowing CLIP to understand a wide range of visual concepts and their linguistic associations. As a result, CLIP demonstrates an impressive ability to generalize across various tasks without the need for additional fine-tuning or training. This zero-shot capability makes CLIP exceptionally versatile and robust, capable of handling diverse and novel classification tasks that conventional models might struggle with. The ability to perform well in a zero-shot scenario significantly reduces the dependency on large annotated datasets, making CLIP a groundbreaking model for applications where data scarcity is a challenge \cite{radford2021learning}.

Each of these models contributes uniquely to the field of image classification. While CNNs like ResNet-32 offer depth and efficiency, transformer-based models like ViT introduce a new paradigm of processing images through global attention mechanisms. Hybrid approaches like CLIP further advance the field by integrating the strengths of both visual and language models, emphasizing the importance of multimodal learning. These developments set the stage for more sophisticated and accurate image classification systems, underscoring the rapid evolution of machine learning techniques in computer vision.

Distribution learning, a fundamental unsupervised learning task, involves both density estimation and generative modeling, serving to unveil the inherent probability distribution of data. Through this task, algorithms aim to capture the intricate patterns and structures within datasets without explicit labels. Employing techniques such as kernel density estimation and parametric models like Gaussian mixture models, algorithms strive to encapsulate the essence of data distributions. Z. Tan et. al utilizes a network architecture and pipeline tailored for distribution learning, specifically for analyzing image pixel distributions using histograms \cite{tan2023affinetransformationinvariant}. This approach proves advantageous when confronted with complex or unknown data distributions, enhancing the flexibility and utility of the analysis.

\section{Proposed Method}

The objective of this project is to classify six different types of garbage: cardboard, glass, paper, metal, trash, and plastic, using convolutional neural networks (CNNs). 

\subsection{Dataset}
\begin{figure}[ht!]
    \centering
    \includegraphics[width=0.5\textwidth]{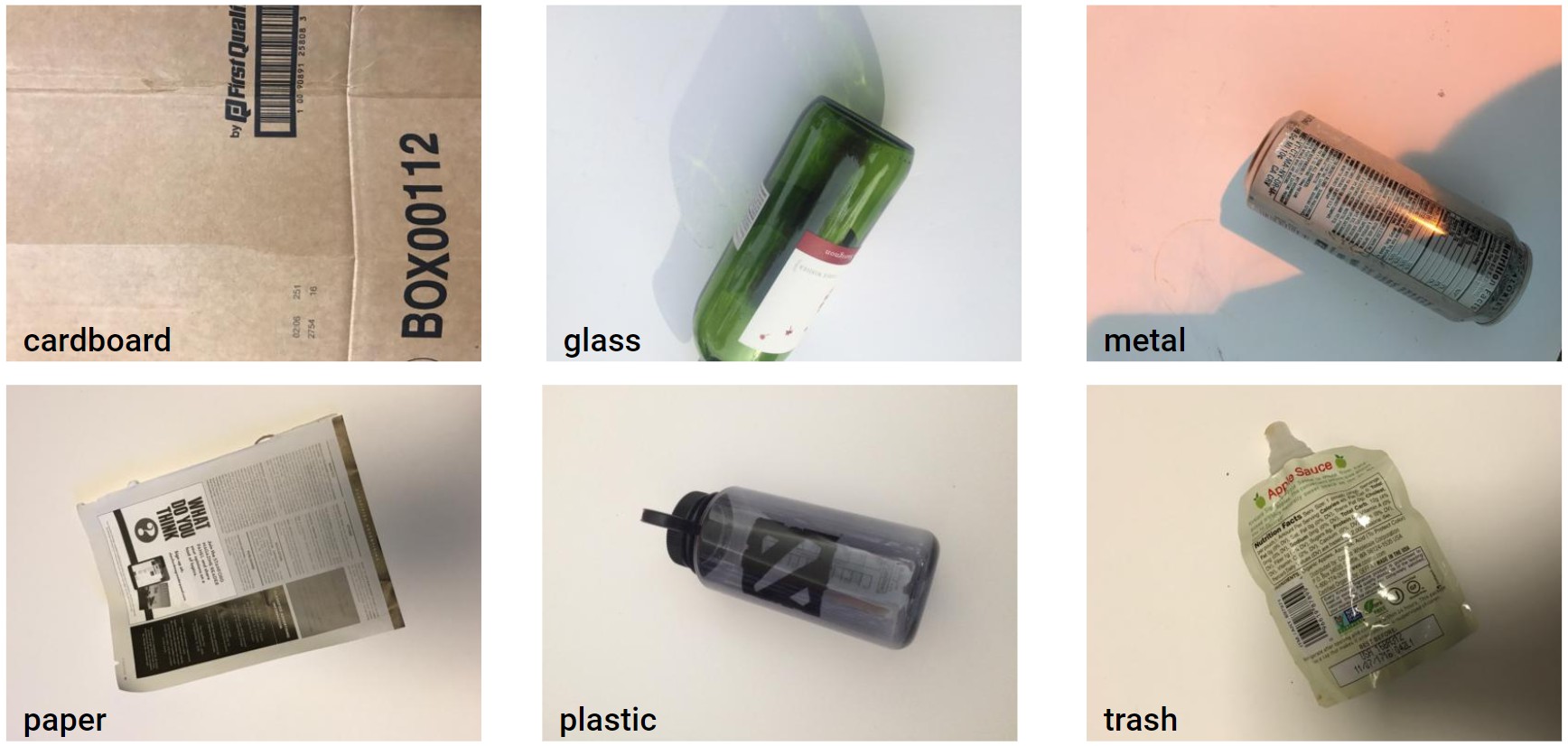}
    \caption{Sample images of each class in the Kaggle Garbage Classification dataset}
    \label{fig:kaggle_example}
\end{figure}

The dataset consists of images categorized into six classes: cardboard, glass, paper, metal, trash, and plastic. The images are resized to a standard size of $224 \times 224$ pixels to maintain uniformity and facilitate processing.\textbf{Figure \ref{fig:kaggle_example}} shows sample images of each class from the dataset.

\subsection{Model Architecture}
The base CNN architecture used for all three experiments is inspired by the VGGNet design principles, consisting of multiple convolutional layers followed by max-pooling layers and fully connected dense layers. The detailed architecture is shown in Table  \ref{tab:model_summary}

\begin{table}[H]
    \centering
    \caption{Model Summary}
    \label{tab:model_summary}
    \begin{tabular}{|l|c|r|}
        \hline
        \textbf{Layer (type)} & \textbf{Output Shape} & \textbf{Param \#} \\
        \hline
        conv2d (Conv2D) & (None, 224, 224, 32) & 896 \\
        \hline
        batch\_normalization (BatchNormalization) & (None, 224, 224, 32) & 128 \\
        \hline
        max\_pooling2d (MaxPooling2D) & (None, 112, 112, 32) & 0 \\
        \hline
        batch\_normalization\_1 (BatchNormalization) & (None, 112, 112, 32) & 128 \\
        \hline
        conv2d\_1 (Conv2D) & (None, 112, 112, 32) & 9,248 \\
        \hline
        batch\_normalization\_2 (BatchNormalization) & (None, 112, 112, 32) & 128 \\
        \hline
        max\_pooling2d\_1 (MaxPooling2D) & (None, 56, 56, 32) & 0 \\
        \hline
        batch\_normalization\_3 (BatchNormalization) & (None, 56, 56, 32) & 128 \\
        \hline
        conv2d\_2 (Conv2D) & (None, 56, 56, 64) & 18,496 \\
        \hline
        batch\_normalization\_4 (BatchNormalization) & (None, 56, 56, 64) & 256 \\
        \hline
        max\_pooling2d\_2 (MaxPooling2D) & (None, 28, 28, 64) & 0 \\
        \hline
        conv2d\_3 (Conv2D) & (None, 28, 28, 64) & 36,928 \\
        \hline
        batch\_normalization\_5 (BatchNormalization) & (None, 28, 28, 64) & 256 \\
        \hline
        max\_pooling2d\_3 (MaxPooling2D) & (None, 14, 14, 64) & 0 \\
        \hline
        conv2d\_4 (Conv2D) & (None, 14, 14, 128) & 73,856 \\
        \hline
        batch\_normalization\_6 (BatchNormalization) & (None, 14, 14, 128) & 512 \\
        \hline
        max\_pooling2d\_4 (MaxPooling2D) & (None, 7, 7, 128) & 0 \\
        \hline
        conv2d\_5 (Conv2D) & (None, 7, 7, 128) & 147,584 \\
        \hline
        batch\_normalization\_7 (BatchNormalization) & (None, 7, 7, 128) & 512 \\
        \hline
        max\_pooling2d\_5 (MaxPooling2D) & (None, 3, 3, 128) & 0 \\
        \hline
        flatten (Flatten) & (None, 1152) & 0 \\
        \hline
        dense (Dense) & (None, 128) & 147,584 \\
        \hline
        dense\_1 (Dense) & (None, 32) & 4,128 \\
        \hline
        dense\_2 (Dense) & (None, 6) & 198 \\
        \hline
        \textbf{Total params} & & \textbf{440,966} \\
        \hline
        \textbf{Trainable params} & & \textbf{439,942} \\
        \hline
        \textbf{Non-trainable params} & & \textbf{1,024} \\
        \hline
    \end{tabular}
\end{table}

\subsection{Experimental setup}
\begin{figure}[ht!]
    \centering
    \includegraphics[width=\textwidth]{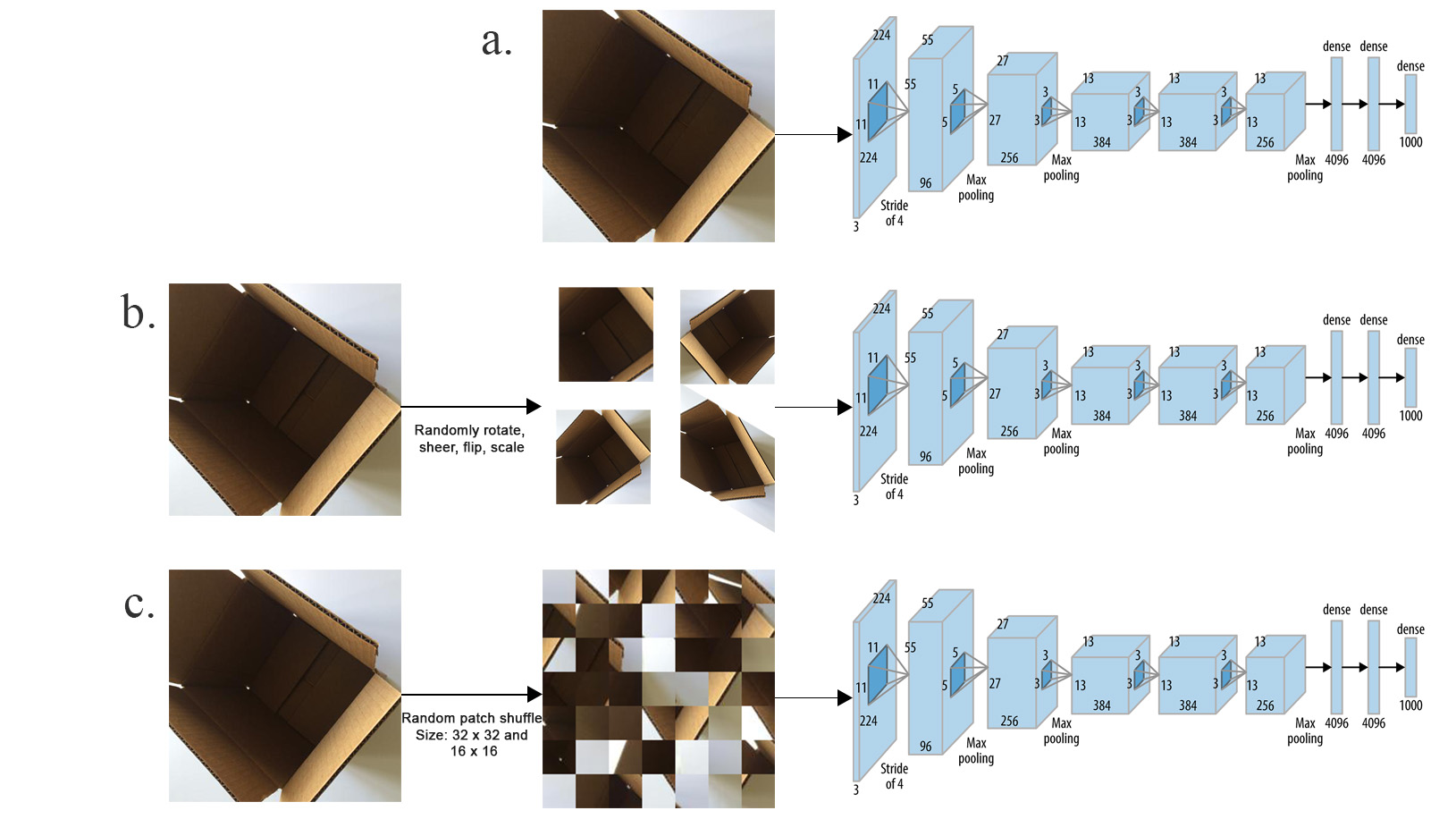}
    \caption{Experimental Pipeline: (a) Original images, (b) Augmented images with random transformations, (c) Images shuffled by patches of size $4 \times 4$ and $32 \times 32$.}
    \label{fig:experiment_pipeline}
\end{figure}

\subsubsection{Experiment A: Training on Original Images}
In the first experiment, we train the CNN on the original dataset. The images are fed directly into the network without any modifications. This serves as the baseline for evaluating the effects of the transformations applied in the subsequent experiments.

\subsubsection{Experiment B: Training on Augmented Images}
In the second experiment, we apply data augmentation techniques to the original images to increase the diversity of the training data. The augmentation operations include random rotations, shear transformations, horizontal and vertical flips, and scaling. This is done to test if the model can generalize better by learning from a more varied set of images.

\subsubsection{Experiment C: Training on Shuffled Patch Images}
In the third experiment, we create two new datasets by randomly shuffling the original images in patches of size $4 \times 4$ and $32 \times 32$ pixels. The purpose of this experiment is to determine if the model can still learn relevant features and achieve good classification performance even when the spatial coherence of the images is disrupted.

\begin{figure}[H]
    \centering
    \includegraphics[width=\textwidth]{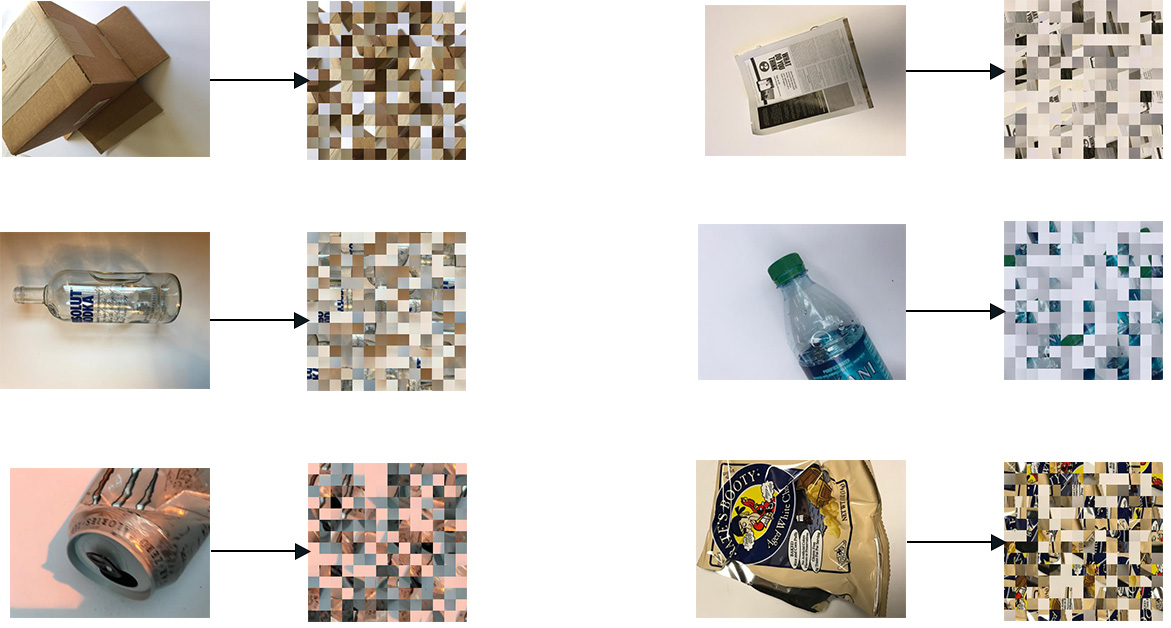}
    \caption{Comparison of Original and Shuffled Images for Garbage Classification}
    \label{fig:classification_results}
\end{figure}

The process of shuffling image patches can be described mathematically as follows:

Let \( I \) be an image of size \( H \times W \times C \), where \( H \) is the height, \( W \) is the width, and \( C \) is the number of channels (e.g., RGB channels).
We divide \( I \) into non-overlapping patches of size \( P \times P \).

\textbf{1. Patch Extraction:}

For each patch:

\[ I_{\text{patch}}(i,j) = I[iP : (i+1)P, jP : (j+1)P] \]

where \( i \) and \( j \) denote the row and column indices of the patch, respectively.

\textbf{2. Random Shuffling:} 

Shuffle the extracted patches using a random permutation:

\[ shuffled\_patches = \text{random\_shuffle}(I_{\text{patch}}) \]

\textbf{3. Reconstruction:} 

Reconstruct the shuffled image from the shuffled patches:

\[ I_{\text{shuffled}}(iP : (i+1)P, jP : (j+1)P) = shuffled\_patches[k] \]

where \( k \) is the index of the current patch in the shuffled list.

To further illustrate the classification performance, Figure \ref{fig:classification_results} shows sample results from each experiment. Each row represents an experiment, showcasing the model's predictions on test images.

\begin{figure}[H]
    \centering
    \includegraphics[width=0.5\textwidth]{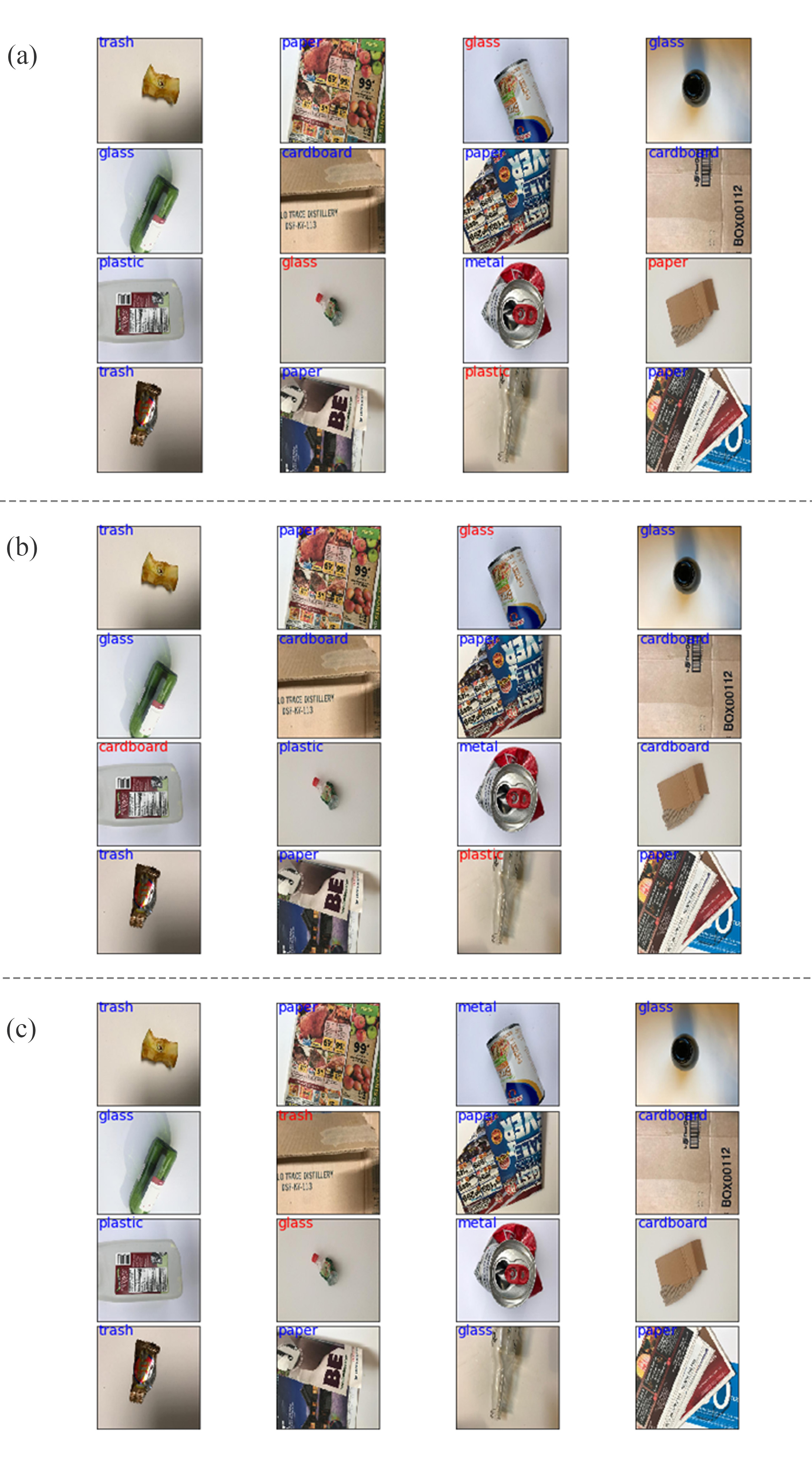}
    \caption{Classification results of the model trained on (a) original images, (b) augmented images, and (c) shuffled images}
    \label{fig:classification_results}
\end{figure}

\section{Results and Discussion}

The objective of this study was to evaluate the performance of different image classification models on garbage classification using various data preprocessing techniques. The results of the validation accuracy for different models and dataset types are summarized in Table~\ref{tab:val_results}.

\begin{table}[htbp]
    \centering
    \small % Adjust font size if necessary
    \caption{Validation Accuracy for Different Models and Dataset Types}
    \label{tab:val_results}
    \begin{tabular}{|l|c|c|c|c|c|}
        \hline
        \textbf{Models trained on vs Testing Dataset Types} & \textbf{Original} & \textbf{4x4 Patch} & \textbf{32x32 Patch} & \textbf{Flipped} & \textbf{Scaled} \\
        \hline
        Original Images & 0.7600 & 0.3560 & 0.4280 & 0.6160 & 0.7320 \\
        \hline
        Augmented Images & 0.7840 & 0.3160 & 0.4280 & 0.6600 & 0.7680 \\
        \hline
        4x4 Shuffled Patch Images & 0.8240 & 0.4530 & 0.5520 & 0.7360 & 0.7480 \\
        \hline
        32x32 Shuffled Patch Images & 0.6600 & 0.4280 & 0.4560 & 0.3160 & 0.3000 \\
        \hline
    \end{tabular}
\end{table}

\subsection{Baseline Model Performance}
The model trained on the original dataset for 200 epochs achieved a validation accuracy of 76\%. This serves as a baseline for comparison with models trained on augmented and shuffled datasets.

\subsection{Data Augmentation Techniques}
The application of data augmentation techniques, including random rotation, shear, flip, and scale, resulted in an improvement in model performance. The augmented images model achieved a validation accuracy of 78.4\% on the original dataset, which is a slight increase compared to the baseline. Additionally, the augmented images model showed improved robustness with a validation accuracy of 76.8\% on the scaled dataset. This demonstrates the effectiveness of data augmentation in enhancing the model's ability to generalize to unseen data.

\subsection{Patch Shuffling Techniques}
Patch shuffling was applied at two different scales: 4x4 and 32x32. The model trained on 4x4 shuffled patch images demonstrated superior performance across all dataset types, achieving the highest validation accuracy of 82.4\% on the original dataset and 73.6\% on the flipped dataset. This suggests that smaller patch shuffling may help the model capture the distribution patterns more effectively, thus improving its generalization capability.

Conversely, the model trained on 32x32 shuffled patch images exhibited a significant drop in performance. It achieved a validation accuracy of only 66\% on the original dataset and 30\% on the scaled dataset. This indicates that larger patch shuffling disrupts the spatial coherence of the images to a greater extent, which adversely affects the model's learning process.

The patch shuffling results indicate that the model learns pixel distribution patterns. This learning is more effective with smaller patch sizes (4x4), which likely preserves essential local spatial information, whereas larger patches (32x32) may obscure such patterns due to excessive disruption.

\subsection{Discussion}
The results suggest that while data augmentation techniques enhance model performance and robustness, the scale of patch shuffling plays a critical role in determining its effectiveness. Small-scale shuffling (4x4 patches) appears to introduce beneficial variations that improve model accuracy, whereas large-scale shuffling (32x32 patches) disrupts essential spatial relationships, leading to poorer performance.

In conclusion, this study highlights the importance of carefully selecting data preprocessing techniques. Data augmentation proved to be beneficial, and small-scale patch shuffling offered additional improvements. However, large-scale patch shuffling should be approached with caution due to its potential to degrade model performance. Future work may explore other forms of data preprocessing and augmentation to further optimize garbage classification models.

\bibliographystyle{plain}
\bibliography{refs}
\end{document}